\documentclass[journal]{IEEEtran}

\usepackage{amsmath,amsfonts}
\usepackage{algorithmic}
\usepackage{algorithm}
\usepackage{array}
\usepackage[caption=false,font=normalsize,labelfont=sf,textfont=sf]{subfig}
\usepackage{textcomp}
\usepackage{stfloats}
\usepackage{url}
\usepackage{verbatim}
\usepackage{graphicx}
\usepackage{cite}
\usepackage{multirow}
\usepackage{color}
\usepackage{booktabs}
\usepackage{xcolor}
\usepackage{colortbl}
\usepackage{makecell}
\usepackage{diagbox}
\usepackage{xfrac}

\usepackage{amsthm,amssymb}
\usepackage{mathrsfs}

\begin{document}

{
\twocolumn

\title{\textsc{\textbf{StereoFlow:}} Progressive Stereo Matching with StereoDiT and Transition Flow Matching}

\author{Hao Wang, Haoran Geng, Xiaotong Yang, Jing Tang, Songlin Wei, Linlong Lang, Yeying Jin, Zheng Zhu, Zhaoxin Fan\IEEEauthorrefmark{1}, Biao Leng\IEEEauthorrefmark{1}

\thanks{This work was supported by the New Generation Artificial Intelligence-National Science and Technology Major Project (2025ZD0122603). It was also supported by the Postdoctoral Fellowship Program and China Postdoctoral Science Foundation under Grant No. 2024M764093 and Grant No. BX20250485, the Beijing Natural Science Foundation under Grant No. 4254100, and by Beijing Advanced Innovation Center for Future Blockchain and Privacy Computing.}
\thanks{Hao Wang, Linlong Lang and Biao Leng are with School of Computer Science and Engineering, Beihang University, Beijing 100191, China (E-mail: {lenny, langlinlong, lengbiao}@buaa.edu.cn). Zhaoxin Fan is with School of Artificial Intelligence, Beihang University, Beijing 100191, China (E-mail: zhaoxinf@buaa.edu.cn). Haoran Geng is with University of California, Berkeley (e-mail: ghr@berkeley.edu). Songlin Wei is with University of Southern California (e-mail: weisongl@usc.edu). Xiaotong Yang is with Google (e-mail: syangx38@gmail.com). Jing Tang is with State Key Laboratory of Intelligent Manufacturing Equipment and Technology, Huazhong University of Science and Technology, Wuhan 430074, China (e-mail: focusers@163.com). Yeying Jin is with National University of Singapore (e-mail: jinyeying@u.nus.edu). Zheng Zhu is with GigaAI (e-mail: zhengzhu@ieee.org).}
\thanks{Corresponding author: Zhaoxin Fan (zhaoxinf@buaa.edu.cn), Biao Leng (lengbiao@buaa.edu.cn)}}

\markboth{SUBMISSION TO IEEE Transactions}%
{Shell \MakeLowercase{\textit{et al.}}: Bare Demo of IEEEtran.cls for IEEE Journals}


\maketitle

\begin{abstract}
Stereo matching is a fundamental task in 3D reconstruction. 
Despite remarkable advances, the prevailing paradigms formulate stereo matching as a deterministic regression problem, collapsing the multimodal distribution modeling into a single-point estimation.
This formulation suffers from a regression-to-mean bias, frequently struggling with ambiguous regions.
In contrast, we introduce a prior-guided generative framework that integrates deterministic matching regression and generative distribution modeling within a complementary formulation.
Built upon this formulation, we introduce \textsc{\textbf{StereoFlow}} through three key components:
(i) a two-stage progressive cascade matching network that progressively produces multi-resolution stereo conditions with complementary matching cues;
(ii) a pixel diffusion transformer (termed StereoDiT) with a frequency-decoupled architecture for modeling correspondence ambiguity; 
(iii) a few-step flow matching objective (termed Transition Flow Matching) for efficient optimization.
In summary, \textsc{\textbf{StereoFlow}} achieves strong geometric consistency and rich fine-grained details in ill-posed, discontinuous regions and under zero-shot generalization. Extensive experiments demonstrate that the proposed \textsc{\textbf{StereoFlow}} establishes multiple state-of-the-art results across benchmarks, including Scene Flow, KITTI, ETH3D, and Middlebury.
\end{abstract}


\section{Introduction}
\label{intro}
\IEEEPARstart{S}{tereo} matching~\cite{survey1} is a fundamental task in computer vision and 3D reconstruction, aiming to estimate dense pixel-wise disparity from rectified stereo image pairs for recovering metric geometry.
Consequently, it has become essential for numerous downstream applications such as autonomous driving, robotic navigation, and augmented/virtual reality~\cite{kitti12, xrstereo}, thereby drawing sustained interest from both the academia and industry community in recent years.

\begin{figure}[t]
    \centering
    \includegraphics[width=1\columnwidth]{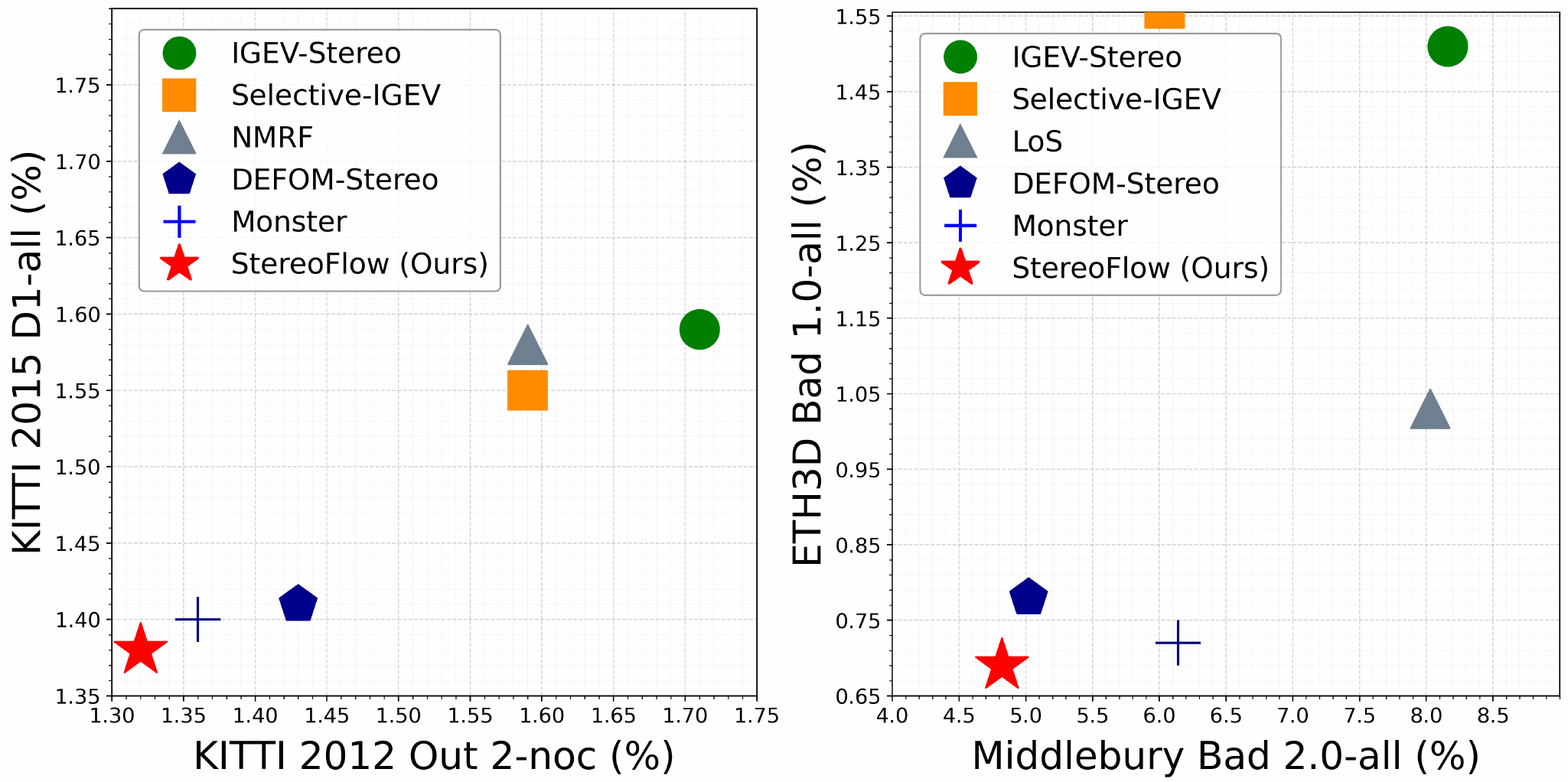}
    \vspace{-18pt}
    \caption{
    \textbf{Comparisons on KITTI-2012~\cite{kitti12}, KITTI-2015~\cite{kitti15}, ETH3D~\cite{eth3d} and Middlebury~\cite{middlebury14}} with the recent state-of-the-art counterparts. The proposed \textsc{\textbf{StereoFlow}} achieves comprehensive performance improvements across various benchmarks.
    (\textit{Zoom in for a better view.})
    }
    \label{fig:benchmark}
\end{figure}
\begin{figure}[t]
    \centering
    \includegraphics[width=1\columnwidth]{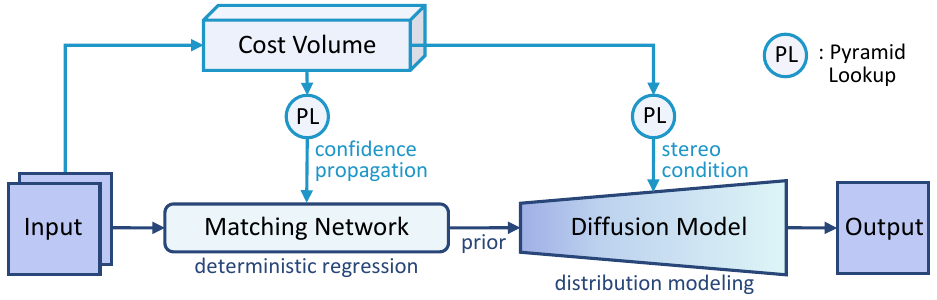}
    \vspace{-18pt}
    \caption{
    \textbf{The prior-guided generative stereo framework} that integrates deterministic matching regression and generative distribution modeling within a complementary formulation, where the former provides disparity priors that substantially constrain the correspondence search space, and the latter explicitly parameterizes multimodal distributions anchored around these priors.
    (\textit{Zoom in for a better view.})
    }
    \vspace{-5pt}
    \label{fig:moti}
\end{figure}
\begin{figure*}[t]
    \centering
    \includegraphics[width=1\linewidth]{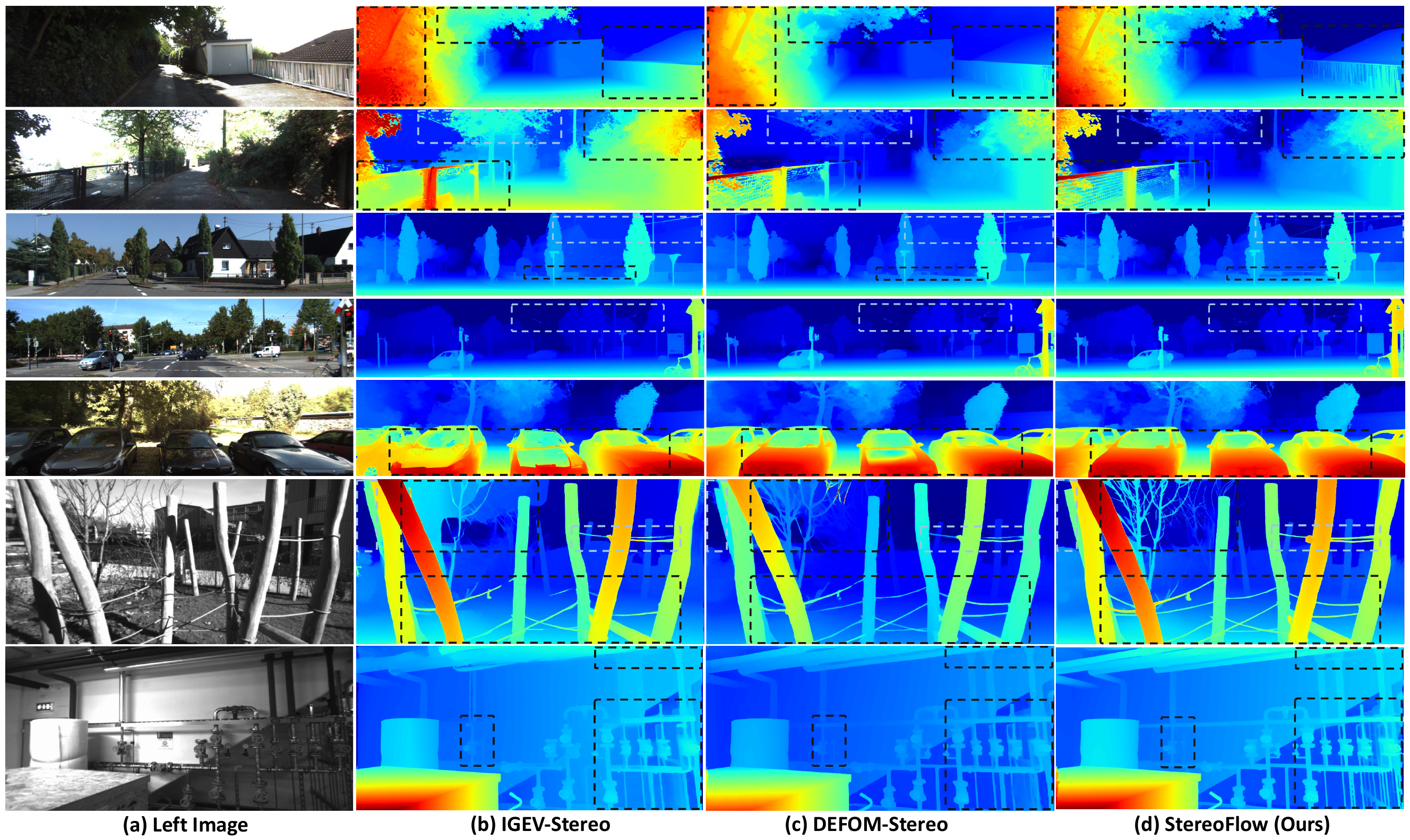}
    \vspace{-18pt}
    \caption{\textbf{Qualitative comparisons of zero-shot generalization} with the baselines IGEV-Stereo~\cite{igev} and DEFOM-Stereo~\cite{defomstereo} on KITTI-2012~\cite{kitti12}, KITTI-2015~\cite{kitti15}, and ETH3D~\cite{eth3d}. All models are trained on the synthetic Scene Flow~\cite{sceneflow} and evaluated directly on the realistic datasets. 
    In contrast, the proposed \textsc{\textbf{StereoFlow}} achieves strong geometric consistency and rich fine-grained details in ambiguous regions such as occlusions, repetitive patterns, reflective surfaces, thin structures, and depth discontinuities under zero-shot generalization.
    (\textit{Zoom in for a better view.})}
    \label{fig:intro}
\end{figure*}

Despite remarkable advances in deep stereo matching~\cite{survey2,survey3}, a fundamental challenge remains in resolving the inherent ambiguity of stereo correspondence.
In the presence of occlusions, repetitive patterns, reflective surfaces, and depth discontinuities, multiple plausible correspondence hypotheses may be equally consistent with the observed stereo evidence~\cite{psmnet,oos,amm}, leading to an intrinsically multimodal correspondence distribution rather than unimodal.
However, the prevailing paradigms~\cite{psmnet,raftstereo,parallax} formulate stereo matching as a deterministic regression problem, collapsing the multimodal distribution modeling into a single-point estimation.
Despite its simplified optimization, this formulation inevitably suffers from a regression-to-mean bias, frequently struggling with ambiguous regions, including: \textbf{(i)} degraded geometric consistency across depth discontinuities, \textbf{(ii)} over-smoothed details in ill-posed regions, and \textbf{(iii)} limited generalization under domain shifts, as illustrated in Fig.~\ref{fig:intro}.

From a probabilistic perspective, stereo matching can be formulated as a problem of distribution modeling conditioned on stereo geometry.
This formulation naturally motivates the adoption of \textit{diffusion models}, which proves exceptionally capable of modeling multimodal distributions.
Existing stereo matching methods typically introduce multimodality only implicitly through loss functions~\cite{psmnet,oos,amm} or uncertainty estimation algorithms~\cite{ucsnet,elf,elfnet}.
In contrast to these deterministic regression models, diffusion models explicitly parameterize multimodal distributions and have achieved significant success across applications such as 3D reconstruction~\cite{latentSplat,diffmatch,diffsplat} and low-level vision~\cite{ddt,deco,sr3,resshift}. 
However, adapting diffusion models to stereo matching remains non-straightforward due to the rigid constraints imposed by epipolar geometry. 
Moreover, without any disparity priors, diffusion models must explore a large solution space along epipolar lines, potentially leading to inefficient dynamics and heavy computational overhead.

In this paper, we observe that deterministic matching regression and generative distribution modeling are complementary: the former  provides disparity priors to constrain the correspondence space, while the latter parameterizes multimodal distributions anchored around these priors.
This formulation regularizes the generative solution space, shifting its focus from exploring the full correspondence manifold to resolving localized ambiguities.
Built upon this formulation, we propose a prior-guided generative stereo framework, as illustrated in Fig.~\ref{fig:moti}, and instantiate it as \textsc{\textbf{StereoFlow}} through three key components:
\textbf{(i)} We introduce a two-stage progressive cascade matching network that produces multi-resolution stereo conditions with complementary matching cues.
The low-resolution stage builds upon a 4D Geometry Encoding Volume (GEV)~\cite{gwcnet,igev,igev++} with geometric consistency, while the high-resolution stage builds upon a 3D All-pairs Correlation Volume (ACV)~\cite{gwcnet,raft,raftstereo} with fine-grained details.
\textbf{(ii)} To model the correspondence ambiguity unresolved by deterministic matching networks, we introduce StereoDiT, a pixel diffusion transformer tailored for stereo matching.
Drawing inspiration from frequency-decomposed disparity representations~\cite{dlnr,mocha,selective}, we formulate StereoDiT as a decoupled architecture:
the geometric encoder enforces low-frequency structural consistency using representations from depth foundation model~\cite{depthanythingv2}, while the stereo decoder recovers high-frequency metric details under specialized stereo conditions.
\textbf{(iii)} Existing diffusion objectives typically construct trajectories from isotropic gaussian priors, failing to utilize the disparity priors and resulting in inefficient dynamics.
To address this limitation, we introduce Transition Flow Matching, a few-step flow matching objective that replaces isotropic gaussian priors with disparity priors.
Further, we formulate the learning dynamics with a linear interpolant for directional displacement of geometric structures and a linear noise scheduler to synthesize details.
This decomposition separates deterministic geometric transport from stochastic perturbations, enabling efficient optimization and few-step sampling.

In summary, \textsc{\textbf{StereoFlow}} achieves strong geometric consistency and rich fine-grained details in ill-posed, discontinuous regions and under zero-shot generalization, as illustrated in Fig.~\ref{fig:intro}.
The proposed \textsc{\textbf{StereoFlow}} establishes multiple state-of-the-art results on the Scene Flow~\cite{sceneflow}, KITTI~\cite{kitti12,kitti15}, ETH3D~\cite{eth3d}, and Middlebury~\cite{middlebury14} benchmarks, as illustrated in Fig.~\ref{fig:benchmark}. 
Ablation studies validate the effectiveness of each component within our framework.
Our contributions are summarized as follows:
\begin{itemize}
\item We propose \textsc{StereoFlow}, a prior-guided generative framework that integrates deterministic matching regression and generative distribution modeling within a complementary formulation.
\item The proposed \textsc{StereoFlow} is built upon three components: (i) a progressive cascade matching network, (ii) a pixel diffusion transformer termed StereoDiT, and (iii) a few-step flow matching objective termed Transition Flow Matching.
\item Extensive experiments demonstrate that the proposed \textsc{StereoFlow} achieves multiple state-of-the-art results across various benchmarks.
\end{itemize}

\section{Related Work}
\label{related}

\subsection{Deep and Generalized Stereo Matching}
Recent advances in deep stereo matching have led to several dominant paradigms.
\textit{Cost filtering-based methods}~\cite{sceneflow,gcnet,psmnet,gwcnet,pcwnet,acvnet,moblestereo,lightstereo} involve steps of cost construction, cost filtering, and disparity regression.
Owing to the regularization of cost filtering, reliable matches are propagated to ambiguous ones to encode non-local geometries within a 4D Geometry Encoding Volume (GEV)~\cite{gwcnet,igev,igev++}. 
However, this comes with high computation cost, limited scalability to large disparities, and pronounced over-smoothing of details.
\textit{Iterative optimization-based methods}~\cite{raftstereo,dlnr,igev,selective,mocha}  recurrently updates disparity field using cost features retrieved from a 3D All-pairs Correlation Volume (ACV)~\cite{gwcnet,raft,raftstereo}.
Based on its precise pixel correspondences, they excel at recovering fine-grained details, but lacks sufficient non-local context to propagate confidence for ill-posed regions, trapping in local minima and leading to poor generalization~\cite{dsmnet,s2m2,dive}.
\textit{Transformer-based methods}~\cite{parallax,sttr,cstr,gmstereo,elfnet,goat} depart from the construction of fixed-disparity cost volume but revisit the problem from a sequence-to-sequence matching perspective.
They excel in long-range dependencies and large disparities, but struggle with ambiguity in ill-posed regions without explicit cost volumes.
\textit{Cascade-based methods}~\cite{casstereo,cfnet,crestereo,crestereo++} have shown strong  generalization ability under domain shifts and disparity distribution discrepancies.
Generalized stereo matching~\cite{meta,dsmnet,dive} has gained sustained focus for domain generalization. 
To mitigate learning from synthetic artifacts, works emerge like domain-invariant representation~\cite{dsmnet,matching}, structural prior~\cite{edgestereo,segstereo}, self-supervised learning~\cite{croco}, transfer learning~\cite{transfer}, meta learning~\cite{meta}, and contrastive learning~\cite{revisiting}. 
\textit{Vision foundation model-based methods}~\cite{formerstereo,foundationstereo,stereoanywhere,defomstereo,monster,bridgedepth,dive} exploit their robust features and monocular depth to handle ill-posed regions and improve generalization.

Despite substantial progress, these prevailing paradigms formulate stereo matching as a deterministic regression problem, collapsing the intrinsically multimodal correspondence distributions into a single-point estimation.
This formulation suffers from a regression-to-mean bias, frequently struggling with ambiguous regions.
In contrast, we introduce \textsc{\textbf{StereoFlow}}, a prior-guided generative framework that integrates deterministic matching regression and generative distribution modeling within a complementary formulation.

\subsection{Diffusion Application and Architecture}
\textit{Diffusion models}~\cite{sde,fm,stochastic} have achieved significant success across numerous applications, including 3D reconstruction~\cite{depthfm,lotus,ppd,diffmatch,mvdd,diffnerf,dpcd}, low-level vision~\cite{prior,resshift,rddm}, and human synthesis~\cite{diffcap,CoShMDM,CoreEditor}.
In stereo matching, DiffuVolume~\cite{diffuvolume} proposes a diffusion-based filter to suppress redundancy in cost volume, yet delivers only marginal improvements over convolutional baselines, underscoring the inadequacy of straightforward diffusion adaptations.
D3RoMa~\cite{d3roma} employs a diffusion prior together with a left-right consistency loss, yet crucially neglects the explicit geometric constraints within cost volumes.
FlowDiffuser~\cite{flowdiffuser}, DiffMVS~\cite{diffMVS} and DMIO~\cite{rethink} adopt lightweight ConvGRUs as denoising backbone, yet two drawbacks persist: (i) the recurrent formulation of ConvGRU entangles representations across diffusion timesteps, and (ii) the limited capacity restricts the expressive distribution modeling required for ambiguous correspondences.
Concurrently, generic diffusion architecture has undergone a transition shifting from \textit{latent diffusion}~\cite{sd,sd3,flux} to \textit{pixel diffusion}~\cite{jit,pixelflow,pyramidfm}, abandoning the compressed VAE latent space in favor of an end-to-end pixel space.
This transition is primarily driven by the need for higher fidelity and finer details, which is critical for stereo matching, where sub-pixel localization and object boundaries are paramount.

Existing diffusion-based stereo methods have proven that straightforward adaptation is insufficient, exposing a persistent misalignment between generic generative paradigms and task-specific stereo frameworks.
Different from generic generation, stereo matching is tightly constrained by epipolar geometry.
Without any disparity priors, diffusion models must explore a large solution space along epipolar lines, potentially leading to inefficient dynamics and heavy computational overhead.
Thus, we propose to utilize the disparity priors from deterministic matching networks to regularize the generative solution space, shifting its focus from exploring the full correspondence manifold to resolving localized ambiguities.
Moreover, drawing inspiration from frequency-decomposed disparity representations~\cite{dlnr,mocha,selective}, we introduce StereoDiT, a pixel diffusion transformer with a frequency-decoupled architecture tailored for stereo matching.

\subsection{Diffusion and Flow Matching Objective}
\textit{Score-based diffusion models}~\cite{ddpm,ncsn,sde} 
have achieved remarkable success and are supported by comprehensive theories~\cite{edm,edm2,ddim,deis,iddpm,onthe}, which bridges target distributions and source gaussian priors with Stochastic Differential Equations (SDE) and a learned score function. 
Diffusion models benefit from stochastic exploration that facilitates smooth distribution transitions and robust mode coverage.
Nevertheless, two fundamental challenges persist: 
(i) the curved SDE trajectories typically require massive neural network evaluations during sampling, even with probability flow ODE and accelerated samplers. 
(ii) the fixed gaussian priors may be poorly aligned with the target distributions, leading to distorted trajectories and unnecessary transport costs.
In parallel, \textit{Flow Matching}~\cite{fm,reflow,stochastic} has emerged as a powerful alternative for bridging arbitrary target and source distributions~\cite{sd3,flux,crossflow,flowtok} with Ordinary Differential Equations (ODE) and a learned velocity field.
In particular, optimal paths~\cite{multisample,otfm,datadependent,instaflow} construct linear interpolant between coupling samples, yielding straighter trajectories, lower transport costs, and substantially faster sampling than diffusion-based approaches~\cite{fm,ddpm}.
However, the deterministic nature of these optimal paths introduces its own limitations:
without stochastic perturbations~\cite{ncsn,stochastic}, directly regressing velocity fields across high-dimensional space from sparse couplings on low-dimensional data manifolds, can lead to generative artifacts and reduced mode coverage. 

\begin{figure*}[!t]
    \centering
    \includegraphics[width=1\textwidth]{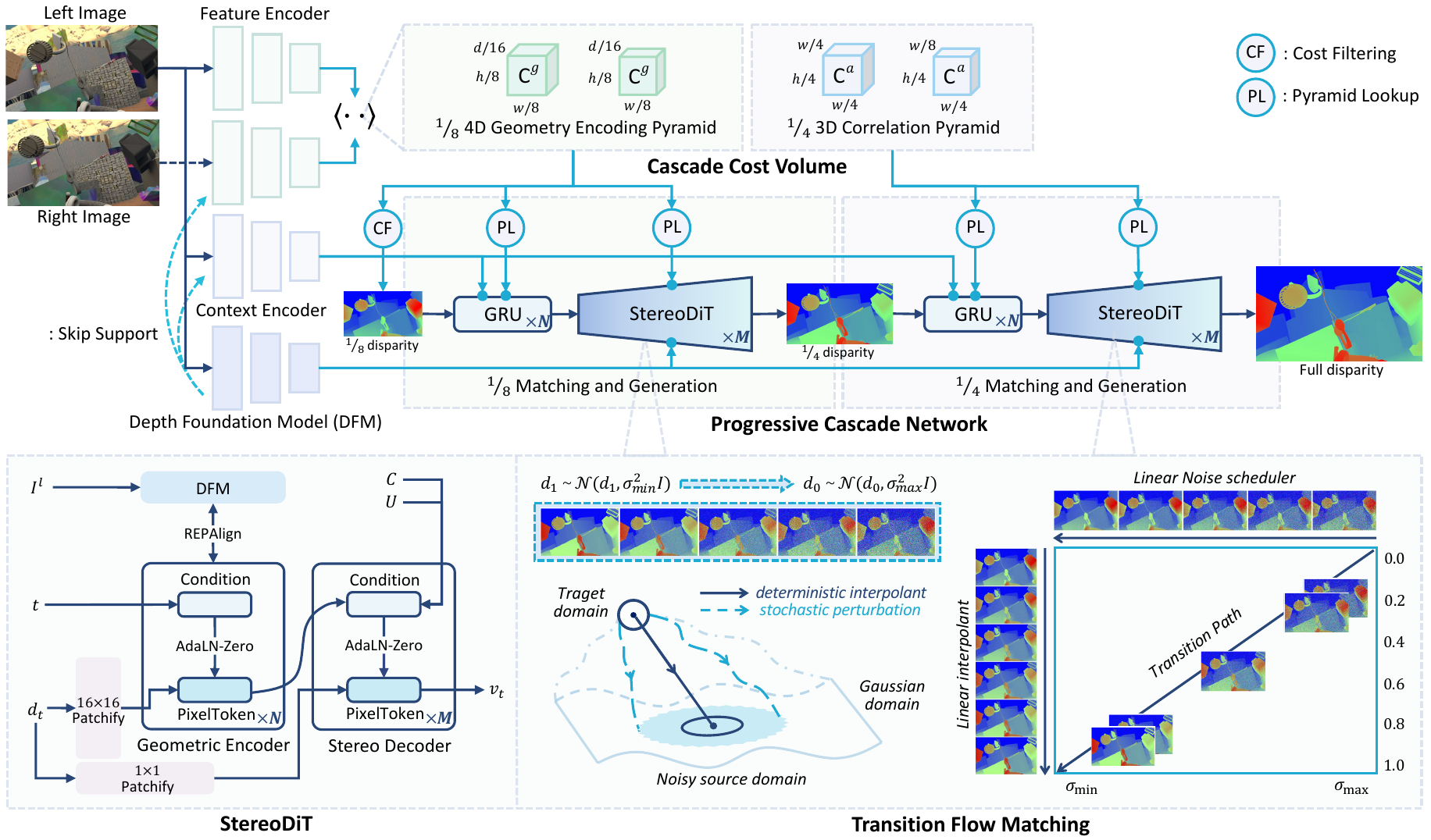}
    \vspace{-18pt}
    \caption{\textsc{\textbf{StereoFlow}} is a prior-guided generative stereo framework built upon three key components: \textbf{(i)} A two-stage progressive cascade matching network that progressively produces multi-resolution stereo conditions with complementary matching cues; 
    \textbf{(ii)} A pixel diffusion transformer (termed StereoDiT) with a decoupled architecture consisting of a geometric encoder for low-frequency structural consistency and a stereo decoder for high-frequency metric details;
    \textbf{(iii)} A few-step flow matching objective (termed Transition Flow Matching) formulates the learning dynamics with a linear interpolant for directional displacement of geometric structures and a linear noise scheduler to synthesize details. (\textit{Zoom in for a better view.})
    }
\label{fig:pipeline}
\end{figure*}

To combine the complementary strengths of diffusion paths and optimal paths, we intrudoce Transition Flow Matching (TFM), a coupling-based flow matching with a linear interpolant for directional displacement of geometric structures and a linear noise scheduler to synthesize details.
By jointly exploiting deterministic transport and stochastic perturbations, this objective preserves the disparity priors inherent in stereo matching, enabling efficient learning dynamics and optimization.

\section{Method}
\subsection{Overview}
The overall architecture of \textsc{\textbf{StereoFlow}} is illustrated in Fig.~\ref{fig:pipeline}.
Given a pair of rectified stereo images $\mathbf{I}^{l,r}$, we first extract multi-level representations using a feature encoder, a context encoder, and a depth foundation model.
These representations are subsequently integrated to construct cascaded cost volumes and iterative states across resolutions.

Built upon these representations, we introduce a two-stage progressive cascade matching network that progressively produces multi-resolution stereo conditions with complementary matching cues.
At each stage, we incorporate a prior-guided generative stereo framework composed of two components: a ConvGRU-based matching network to produce disparity priors, and a StereoDiT-based generative model to parameterize multimodal correspondence distributions anchored around these priors.
Specifically, we formulate StereoDiT as a pixel diffusion transformer with a decoupled architecture consisting of a geometric encoder and a stereo decoder. 
Further, we introduce Transition Flow Matching, a few-step flow matching objective that formulates the learning dynamics with a linear interpolant and a linear noise schedule.

By integrating deterministic matching regression and generative distribution modeling within a complementary formulation, \textsc{\textbf{StereoFlow}} explicitly models multimodal correspondence distributions, thereby resolving matching ambiguities caused by occlusions, repetitive patterns, reflective surfaces, and depth discontinuities.

\subsection{Progressive Cascade Matching Network}
To integrate low-resolution consistent disparity reconstruction and high-resolution fine-grained disparity recovery within a unified framework, we introduce a progressive cascade network: 
the low-resolution stage builds upon a 4D Geometry Encoding Volume (GEV)~\cite{gwcnet,igev,igev++} that encodes non-local geometries and handles ill-posed regions for geometric consistency, while the high-resolution stage builds upon a 3D All-pairs Correlation Volume (ACV)~\cite{gwcnet,raft,raftstereo} that establishes precise pixel correspondences for fine-grained details.

\noindent \textbf{Feature extraction.} 
There are three encoders in feature extraction:
(i) The feature encoder extracts matching features $\mathbf{f}^{l,r}$ from the stereo images pair $\mathbf{I}^{l,r}$ to construct stereo hypothesis volumes under epipolar constraints.
(ii) The context encoder extracts context features $\mathbf{c}^l$ from the left image $\mathbf{I}^l$ to guide confidence propagation.
(iii) The depth foundation model extracts monocular depth features $\mathbf{d}^{l,r}$ from the paired images $\mathbf{I}^{l,r}$ to provide geometrically consistent features.
Furthermore, we integrate the monocular depth features into matching features to alleviate matching ambiguity in ill-posed regions, and into context features to mitigate domain biases from synthetic artifacts.
Given the image pair $\mathbf{I}^{l,r}\in\mathbb{R}^{{3}\times h\times w}$, we construct the multi-scale matching pyramid of $\{\mathbf{f}^{l,r}_s\in\mathbb{R}^{{c_s}\times{\sfrac{h}{2^s}}\times{\sfrac{w}{2^s}}}\}_{s=1}^4$, context pyramid of $\{\mathbf{c}^l_s\in\mathbb{R}^{{c_s}\times{\sfrac{h}{2^s}}\times{\sfrac{w}{2^s}}}\}_{s=1}^3$, and monocular depth pyramid of $\{\mathbf{d}^{l,r}_s\in\mathbb{R}^{{c_s}\times{\sfrac{h}{2^s}}\times{\sfrac{w}{2^s}}}\}_{s=1}^4$ through the above encoders.

\noindent \textbf{Cascade cost volume construction.} 
There are two cost volumes in cost construction: 
(i) The 4D Geometry Encoding Volume (GEV) $\mathbf{C}^g_{s=3}\in\mathbb{R}^{{c}\times{\sfrac{d}{2^3}}\times{\sfrac{h}{2^3}}\times{\sfrac{w}{2^3}}}$ is a regularized combined cost volume formed by a group-wise correlation volume and a compact concatenation volume, where $c$ denotes disparity channel dimension and $d$ denotes the number of disparity candidates.
Such a volume preserves sufficient contextual information from the matching features, whereas the regularization process propagates reliable matches to ambiguous ones for encoding non-local geometries and handling ill-posed regions.
(ii) The 3D All-pairs Correlation Volume (ACV) $\mathbf{C}^a_{s=2}\in\mathbb{R}^{{\sfrac{w}{2^2}}\times{\sfrac{h}{2^2}}\times{\sfrac{w}{2^2}}}$ is a single-channel correlation volume with full-range disparity candidates.
Such a volume establishes precise pixel correspondence for recovering fine-grained details and handling large disparities.
We build the 4D GEV from low-resolution matching features at $\sfrac{1}{2^3}$ scale, as their wide receptive fields and abstract semantics better handle ill-posed regions. 
We build the 3D ACV from high-resolution matching features at $\sfrac{1}{2^2}$ scale, as their low-level textures better establish reliable pixel correspondence.
The multi-level GEV pyramid of $\{\mathbf{C}^{g}_{s=3,l}\}_{l=1}^2$ and ACV pyramid of $\{\mathbf{C}^a_{s=2,l}\}_{l=1}^2$ are further obtained by applying 1D average pooling along the disparity dimension, halving its size at each level.

\noindent \textbf{Progressive cascade matching network.}
There are two stages in progressive cascade network:
(i) At $\sfrac{1}{2^3}$ scale, the initial disparity is first regressed from the 4D GEV $\mathbf{C}^g_{s=3}$ via the cost filtering of a lightweight 3D network.
Then, the disparity field is iteratively optimized using cost features retrieved from the 4D GEV pyramid $\{\mathbf{C}^{g}_{s=3,l}\}_{l=1}^2$ via pyramid lookup operation.
This stage capitalizes on the strengths of low-resolution GEV and low-resolution features for consistent disparity reconstruction.
(ii) At $\sfrac{1}{2^2}$ scale, the upsampled disparity field is iteratively optimized using correlation features retrieved from the 3D ACV pyramid $\{\mathbf{C}^a_{s=2,l}\}_{l=1}^2$ via pyramid lookup operation.
This stage capitalizes on the strengths of high-resolution ACV and high-resolution features for fine-grained and large disparity recovery.
In contrast to prior cascade networks~\cite{casstereo,cfnet,crestereo}, we claim distinct roles of each stage combined with complementary stereo geometries, forming a core architectural contribution of our work. 

\noindent \textbf{Prior-guided generative stereo framework.}
At each stage, we employ a prior-guided generative stereo framework comprising two components: 
(i) a ConvGRU-based matching network to produce disparity priors, and (ii) a StereoDiT-based generative model to parameterize multimodal correspondence distributions anchored around these priors.

\subsubsection{\textbf{ConvGRU-based matching regression}}
Following popular RAFT-style optimization~\cite{raft,raftstereo}, the ConvGRU-based matching regression recurrently updates disparity fields ${d}_{n}$ as follows: 
\begin{equation}
\begin{aligned}
{z}_n = &~ \text{Conv}({d}_{n-1}, {C}_{n-1}), \\
\Delta{{d}} = &~ \text{ConvGRU}({h}_{n-1}, {z}_n, {c}), \\
{d}_{n}= &~ {d}_{n-1}+\Delta{{d}}. 
\end{aligned}
\label{eq:gru}
\end{equation}
where ${C}_{n-1}$ denotes the local cost features indexed by ${d}_{n-1}$ and retrieved from cost volumes $\mathbf{C}$, ${z}_n$ denotes the motion features, ${h}_{n-1}$ denotes the hidden state from previous iteration, and ${c}$ denotes the context features. 
This component suffers from two critical limitations:
(i) its intrinsic locality restricts confidence propagation for ambiguous regions, trapping it in local minima and resulting in weak generalization~\cite{dsmnet,dive,s2m2}. 
(ii) its deterministic matching paradigm restricts modeling of multi-modal distributions inherent in ambiguous regions.

\subsubsection{\textbf{StereoDiT-based distribution modeling}}
Given the updated disparity $d$ as prior, the StereoDiT-based distribution modeling is introduced to parameterize multimodal correspondence distributions anchored around these priors, as follows:
\begin{equation}
\begin{aligned}
{d} &,{C},{{d}_*}= \text{Up}({d}, {C}, {{d}_*}), \\
{d} &= \text{StereoDiT}({d}, {C}, {{d}_*}). \\
\end{aligned}
\label{eq:dit}
\end{equation}
where \text{Up} denotes the convex upsample, ${C}$ denotes the local cost features indexed by ${d}$ and retrieved from cost volumes $\mathbf{C}$, and ${{d}_*}$ denotes the monocular depth features. 
This component models the correspondence ambiguity unresolved by deterministic matching networks, enabling robust reconstruction in ambiguous regions.

\subsection{StereoDiT}
StereoDiT is a pixel diffusion transformer featuring a decoupled architecture of a geometric encoder and a stereo decoder. 
The geometric encoder employs monocular depth features to enforce low-frequency geometric consistency, while the stereo decoder employs a set of specialized stereo conditions to recover high-frequency metric details.

\noindent \textbf{Geometric encoder.} The encoder consists of multiple stacked diffusion transformer layers with Attention, FFN, RoPE and AdaLN-Zero, under a large patch size of $16^2$. Given the noisy disparity ${d}_t$ and timestep $t$ as inputs, it outputs the self‑condition feature ${z}_t$.
\begin{equation}
    {z}_t = \text{Encoder}_{16^2}({d}_t, t).
    \label{eq:enc}
\end{equation}
To further incorporate geometric consistency from monocular depth features to ${z}_t$, we employ the REPresentation Alignment (REPA) technique~\cite{repa,repae} that aligns the intermediate features ${h}_i$ from the $i$-th layer of encoder with monocular depth features ${d}_*$. 
\begin{equation}
    \mathcal{L}_{\text{repa}} = 1-\text{cos}(h_{\phi}({h}_i), {d}_*).
    \label{eq:repa}
\end{equation}
where $h_{\phi}$ denotes a learnable projection MLP, and $\text{cos}$ is the cosine similarity to measure the similarity between features. 
It has been widely proven that the learning dynamics can be more efficient by introducing external representations~\cite{repa,repae,ppd}.

\noindent \textbf{Stereo decoder.} The decoder consists of multiple lightweight MLPs under a small patch size of $1^2$.
Given the noisy disparity ${d}_t$, timestep $t$, self-condition feature ${z}_t$, local cost features ${C}$ and uncertainty ${U}$ (warping the right image to the left and calculate the uncertainty.) as inputs, it outputs the velocity ${v}_t$:
\begin{equation}
    {v}_t = \text{Decoder}_{1^2}({d}_t, t, {z}_t, {C}, {U}).
    \label{eq:dec}
\end{equation}
The self-condition features $z_t$ facilitate geometric consistency, the local cost features $C$ enforces stereo epipolar constraints, and the uncertainty $U$ promotes high-frequency details.

\subsection{Transition Flow Matching}
Transition Flow Matching is formulated as a few-step flow matching objective with a linear interpolant for directional displacement of geometric structures and a linear noise scheduler to synthesize details.
This decomposition separates deterministic geometric transport from stochastic perturbations, enabling efficient learning dynamics and optimization.

\noindent \textbf{Transition path.} Given the coupling pair of target-degraded disparities $(d_1, d_0)$, we define the disparity reconstruction as a deterministic transport between the target disparity $d_1$ and the degraded disparity $d_0$. 
In the proposed transition flow matching, the transition path between the target disparity distribution $p(d_1)$ and degraded disparity distribution $p(d_0)$ is formulated as follows:
\begin{equation}
\begin{aligned}
    p_t(d|d_0,d_1)=&~ \mathcal{N}(d|\mu_t(d_0,d_1),\sigma_t(d_0,d_1)^2I), \\
    \mu_t(d_0,d_1)=&~ td_1+(1-t)d_0, \\
    \sigma_t(d_0,d_1)=&~ t\sigma_{\min}+(1-t)\sigma_{\max}.
\end{aligned}
\end{equation}
where $\mu_t$ denotes the linear interpolant, and $\sigma_t$ denotes the linear noise scheduler.
Along the path, the target disparity distribution $p(d_1)$ is modeled as $\mathcal{N}(d_1, \sigma_{\min}^2I)$ with a minimum smoothing factor $\sigma_{\min}=0.01$ to avoid singularity, while the noise-carrying degraded disparity distribution $p(d_0)$ follows $\mathcal{N}(d_0, \sigma_{\max}^2I)$ with a maximum perturbation factor $\sigma_{\max}=0.2$ to ensure generativity and generalization.

\noindent \textbf{Transition flow.} Among the infinite possible vector fields that generate the transition path, we consider the simplest affine transformation flow $\psi_t(d|d_0,d_1)$ and its corresponding vector field $u_t(d|d_0,d_1)$ as follows:
\begin{equation}
\begin{aligned}
    \psi_t(d|d_0,d_1)=&~ \sigma_t(d_0,d_1)\epsilon + \mu_t(d_0,d_1), \\
    u_t(d|d_0,d_1)=&~ \frac{d}{dt}\sigma_t(d_0,d_1)\epsilon + \frac{d}{dt}\mu_t(d_0,d_1). \\
\end{aligned}
\end{equation}

\noindent \textbf{Transition flow matching loss.} In this case, the transition flow matching loss takes the form:
\begin{equation}
    \mathcal{L}_{\text{tfm}} = \mathbb{E}_{t,d_0,d_1}\| v_t(\psi_t(d|d_0,d_1))-u_t(d|d_0,d_1) \|^2.
    \label{eq:fm}
\end{equation}
where $t\sim \text{U}[0,1]$, $d_0\sim p(d_0)$ and $d_1\sim p(d_1)$, and the velocity $v_t$ is parametered by the proposed StereoDiT. 

\subsection{Loss Function}
For supervising deterministic matching regression, we apply a smooth L1 loss to the initial disparity ${d}_{init}$ regressed from the 4D GEV, and an L1 loss to the sequence of iteratively updated disparities $\{{d}_{i}\}_{i=1}^{N}$ produced by the ConvGRU:
\begin{equation}
\begin{aligned}
    \mathcal{L}_{\text{init}} = &~ \text{Smooth}_{L_1}({d}_{init}-{d}_{gt}), \\
    \mathcal{L}_{\text{iter}} = &~ \sum\nolimits_{i=1}^{N} \gamma^{N-i}_{\text{iter}} ||{d}_i-{d}_{gt}||_1, \\
    \mathcal{L}_{\text{match}} = &~ \mathcal{L}_{\text{init}} + \mathcal{L}_{\text{iter}}.
\end{aligned}
\end{equation}
where ${d}_{gt}$ is the ground-truth disparity and $\gamma_{\text{iter}}=0.9$ governs an exponentially increasing weighting scheme over iterations.

For supervising generative distribution modeling, we compute the REPA regularization loss on the intermediate features of StereoDiT, and the Transition Flow Matching objective on the predicted velocity field:
\begin{equation}
    \mathcal{L}_{\text{generative}} = \gamma_{\text{repa}}\mathcal{L}_{\text{repa}} + \mathcal{L}_{\text{tfm}}. \\
\end{equation}
where $\gamma_{\text{repa}}=0.5$ is the weight coefficient for the REPA loss.

The overall objective is formulated as a weighted sum of the matching and generative losses:
\begin{equation}
    \mathcal{L}_{\text{total}} = \gamma_{\text{match}}\mathcal{L}_{\text{match}} + \mathcal{L}_{\text{generative}}. \\
\end{equation}
where $\gamma_{\text{iter}}=0.1$ defines the weight coefficient for the matching loss.

\section{Experiments}
We compare the proposed \textsc{\textbf{StereoFlow}} with the recent SoTA counterparts on various benchmarks. \textsc{\textbf{StereoFlow}} consistently outperforms the recent SoTA counterparts~\cite{igev,defomstereo,monster,bridgedepth}, as illustrated in Fig. \ref{fig:benchmark}.
\begin{figure*}[!t]
    \centering
    \includegraphics[width=1\textwidth]{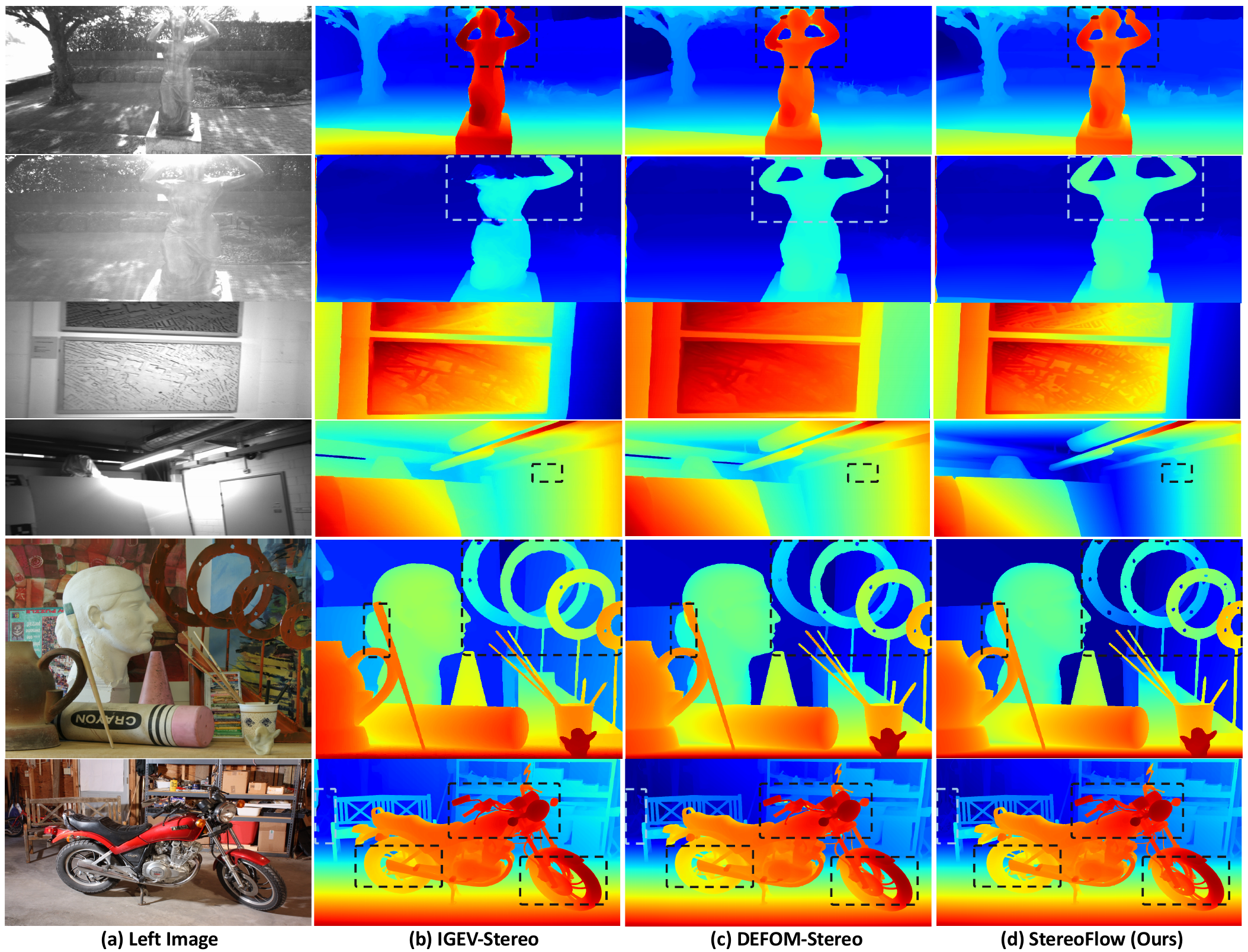}
    \vspace{-20pt}
    \caption{\textbf{Qualitative comparisons on ETH3D and Middlebury} with the baselines IGEV-Stereo~\cite{igev} and DEFOM-Stereo~\cite{defomstereo}. All models are finetuned on the realistic ETH3D~\cite{eth3d} and Middlebury~\cite{middlebury14} datasets. 
    In contrast, the proposed \textsc{\textbf{StereoFlow}} achieves strong geometric consistency and fine-grained details across challenging ill-posed regions, including reflective surfaces, highlights, textureless, occluded and discontinuous areas.    
    (\textit{Zoom in for a better view.})
    }
\label{fig:exp}
\end{figure*}

\subsection{Implementation Details}
\noindent \textbf{Datasets.} Following stardand practice~\cite{defomstereo,monster}, we evaluate the proposed \textsc{\textbf{StereoFlow}} across datasets including Scene Flow~\cite{sceneflow}, KITTI~\cite{kitti12,kitti15}, ETH3D~\cite{eth3d} and Middlebury~\cite{middlebury14}. 
For finetuning on the ETH3D and Middlebury, we construct the Basic Training Set (BTS), a mixture of datasets, including Scene Flow~\cite{sceneflow}, CREStereo~\cite{crestereo}, Tartan Air~\cite{tartanair}, Sintel Stereo~\cite{sintel}, FallingThings \cite{fallingthings3d} and InStereo2k~\cite{instereo2k}.

\noindent \textbf{Baselines.} 
We implement the proposed \textsc{\textbf{StereoFlow}} using PyTorch on NVIDIA A100 GPUs.
Our baseline is a simplified version of DEFOM-Stereo~\cite{defomstereo}, with its affine-invariant monocular depth and metric scaling modules removed.
Following previous works~\cite{defomstereo,monster,bridgedepth}, we adopt DepthAnythingV2-L~\cite{depthanythingv2} as our depth foundation model. 
We evaluate the proposed model against the recent SoTA counterparts, including Selective-IGEV~\cite{selective}, NMRF~\cite{nmrf}, Mocha-Stereo~\cite{mocha}, DEFOM-Stereo~\cite{defomstereo}, Monster~\cite{monster}, BridgeDepth~\cite{bridgedepth}, and FoundationStereo~\cite{foundationstereo}, among others.

\noindent \textbf{Training details.}
We pre-train \textsc{\textbf{StereoFlow}} on the synthetic Scene Flow dataset, followed by fine-tuning on realistic datasets. 
The pre-training process consists of two stages:: 
(i) We first train the cascade matching network without StereoDiT for 200k steps, using the matching loss $\mathcal{L}_{\text{match}}$, the AdamW optimizer, a one-cycle learning rate of 2e-4, and a batch size of 16.
(ii) We then train the full model — comprising both the matching network and StereoDiTs — for an additional 100k steps, using the overall loss $\mathcal{L}_{\text{total}}$, the AdamW optimizer, a one-cycle learning rate of 1e-4, and a batch size of 16.
For fine-tuning on the realistic datasets, we train the full model using the overall loss $\mathcal{L}_{\text{total}}$, the AdamW optimizer, a one-cycle learning rate of 1e-4, and a batch size of 8.

\subsection{Benchmark Evaluation}
\label{sec:synthetic}
\setlength{\tabcolsep}{12pt}
\begin{table*}[t]
    \caption{\textbf{Synthetic pre-training evaluation on Scene Flow test set.} The proposed \textsc{StereoFlow} achieves SoTA performance against the recent counterparts.}
    \vspace{-6pt}
    \centering
    \resizebox{1\textwidth}{!}{
    \begin{tabular}{l|cccccc|c}
    \toprule
    \textbf{Methods} & \makecell{RAFT-Stereo\\ \cite{raftstereo}} & \makecell{IGEV-Stereo\\ \cite{igev}} & \makecell{NMRF\\ \cite{nmrf}} & \makecell{Selective-IGEV\\ \cite{selective}} &  \makecell{DEFOM-Stereo\\ \cite{defomstereo}} & \makecell{Monster\\ \cite{monster}} & \makecell{\textbf{\textsc{StereoFlow}}\\ (Ours)} \\
    \midrule
    EPE [px] $\downarrow$ & 0.56 & 0.47 & 0.45 & 0.44 & 0.42 & 0.38 & \textbf{0.32} \\
    \bottomrule
    \end{tabular}
    }
    \vspace{-2pt}
    \label{tab:sceneflow}
\end{table*}
\setlength{\tabcolsep}{6pt}
\begin{table*}[t]
    \caption{\textbf{Synthetic pre-training evaluation on Scene Flow test set.} The proposed \textsc{StereoFlow} achieves SoTA performance against the recent counterparts.}
    \vspace{-6pt}
    \centering
    \resizebox{1\textwidth}{!}{
    \begin{tabular}{l|ccc|ccc|cccc|cccc}
     \toprule
     \multirow{3}{*}{\textbf{Methods}} & \multicolumn{3}{c|}{\textbf{ETH3D}} & \multicolumn{3}{c|}{\textbf{Middlebury}} & \multicolumn{4}{c|}{\textbf{KITTI-2015}} & \multicolumn{4}{c}{\textbf{KITTI-2012}}  \\
     \cline{2-15}
     & Bad1.0 & Bad1.0 & RMSE & Bad2.0 & Bad2.0 & RMSE & D1-bg & D1-all & D1-bg & D1-all & Out-2 & Out-2 & Out-3 & Out-3\\
     & Noc & All & Noc & Noc & All & Noc & Noc  & Noc & All &  All & Noc & All & Noc & All \\
    \midrule
    GwcNet~\cite{gwcnet} & 6.42 & 6.95 & 0.69 & - & - & - & 1.61 & 1.92 & 1.74 & 2.11 & 2.16 & 2.71 & 1.32 & 1.70 \\
    GANet~\cite{ganet}  & 6.22 & 6.86 & 0.75 & - & - & - & 1.40 & 1.73 & 1.55 & 1.93  & 1.89 & 2.50 & 1.19 & 1.60 \\
    ACVNet~\cite{acvnet} & 2.58 & 2.86 & 0.45 & 13.70 & 19.50 & 32.2 & 1.26 & 1.52 & 1.37 & 1.65  & 1.83 & 2.35 & 1.13 & 1.47 \\
    RAFT-Stereo~\cite{raft} & 2.44 & 2.60 & 0.36 & 4.74 & 9.37 & 8.41 & 1.44 & 1.69 & 1.58 & 1.82  & 1.92 & 2.42 & 1.30 & 1.66 \\
    CREStereo~\cite{crestereo} & 0.98 & 1.09 &  0.28 & 3.71 & 8.13 & 7.70 & 1.33 & 1.54 & 1.45 & 1.69  & 1.72 & 2.18 & 1.14 & 1.46 \\
    CroCo~\cite{croco} & 0.99 & 1.14 & 0.30 & 7.29 & 11.11 & 8.91 & 1.30 & 1.51 & 1.38 & 1.59  & - & - & - & - \\
    DLNR~\cite{dlnr} & - & - & - & 3.20 & 6.98 & 7.78 & 1.42 & 1.61 & 1.60 & 1.76  & - & - & - & - \\
    IGEV-Stereo~\cite{igev} & 1.12 & 1.51 & 0.34 & 4.83 & 8.16 & 12.80 & 1.27 & 1.49 & 1.38 & 1.59  & 1.71 & 2.17 & 1.12 & 1.44 \\
    Selective-IGEV~\cite{selective} & 1.23 & 1.56 & 0.29 &  2.51 & 6.04 & 7.26 & 1.22 &  1.44 & 1.33 &  1.55  &  1.59 &  2.05 & 1.07 & 1.38 \\
    LoS~\cite{los} &  0.91 & 1.03 & 0.31 & 4.20 & 8.03 &  6.99 & 1.29 & 1.52 & 1.42 & 1.65  & 1.69 & 2.12 & 1.10 & 1.38 \\
    NMRF~\cite{nmrf} & - & - & - & - & - & - & 1.18 & 1.46 & 1.28 & 1.57 & 1.59 & 2.07 &  1.01 &  1.35 \\
    DEFOM-Stereo~\cite{defomstereo} & 0.70 & 0.78 & 0.22 & \underline{2.39} & \underline{5.02} & \underline{5.81} & 1.15 & 1.33 & 1.25 & 1.41 & 1.43 & 1.79 & 0.94 & 1.18 \\
    MonSter~\cite{monster} & \textbf{0.46} & \underline{0.72} & \underline{0.20} & 2.64 & 6.14 & 6.71 & \underline{1.05} & 1.33 & \textbf{1.13} & 1.41 & 1.36 &  1.75 & 0.84 & 1.09 \\
    BridgeDepth~\cite{bridgedepth} & - & - & - & - & - & - &  \underline{1.05} & \underline{1.31} & \textbf{1.13} & \underline{1.40} & \textbf{1.32} & \textbf{1.65} & \textbf{0.83} &  \textbf{1.03} \\
    \midrule
    \textbf{\textsc{StereoFlow}} (Ours) & \underline{0.58} & \textbf{0.69} & \textbf{0.19} & \textbf{2.30} & \textbf{4.82} & \textbf{5.40} &  \textbf{1.03} & \textbf{1.30} & \textbf{1.13} & \textbf{1.38}  &  \textbf{1.32} & \underline{1.71} & \textbf{0.83} & \underline{1.07} \\
    \bottomrule
    \end{tabular}
    }
    \vspace{-2pt}
    \label{tab:mix}
\end{table*}
\setlength{\tabcolsep}{1pt}
\begin{table}[t]
    \centering
    \caption{\textbf{Zero-shot generalization evaluation.} All models are pre-trained on the synthetic Scene Flow and evaluated directly on the realistic datasets. Standard thresholding error rates are used: 3-pixel for KITTI, 2-pixel for Middlebury, and 1-pixel for ETH3D.}
    \vspace{-5pt}
    \resizebox{0.48\textwidth}{!}{
    \begin{tabular}{l|cc|c|c}
    \toprule
    \multirow{1}{*}{\textbf{Methods}} & \textbf{KITTI-2012} & \textbf{KITTI-2015}  & \textbf{Middlebury-Quarter} & \textbf{ETH3D} \\
    \midrule
    DSMNet~\cite{dsmnet} & 6.2 & 6.5 & 8.1 & 6.2 \\
    RAFT-Stereo~\cite{raft} & 4.3 & {5.7} & {9.3} & {3.2} \\ 
    DLNR~\cite{dlnr} & 9.0 & 16.0 & 7.8 & 22.9 \\
    IGEV-Stereo~\cite{igev}  & 5.1 & 6.0 & 8.8 & 4.0 \\
    Selective-IGEV~\cite{selective} & 5.6 & 6.0 & 9.8 & 6.0 \\
    NMRF~\cite{nmrf} & 4.2 & 5.5 & 7.4 & 3.8 \\
    Mocha-Stereo~\cite{mocha}  & 4.8 & 5.9 & 7.3 & 3.8 \\
    DEFOM-Stereo~\cite{defomstereo}  & 3.7 & 4.9 & 5.6 & 2.3 \\
    Monster~\cite{monster}  & \underline{3.6} & \textbf{3.9} & 5.1 & 2.0 \\
    BridgeDepth~\cite{bridgedepth}  & \underline{3.6} & 4.5 & \underline{4.3} & \textbf{1.3} \\
    \midrule
    \textbf{\textsc{StereoFlow}} (Ours) & \textbf{3.5} & \textbf{3.9} & \textbf{4.1} & \underline{1.7} \\
    \bottomrule
  \end{tabular}
    }
    \vspace{-8pt}
    \label{tab:zero-shot}
\end{table}
\noindent \textbf{Synthetic Pre-training and Zero-shot Generalization.}
For pre-training evaluation in Tab.~\ref{tab:sceneflow}, the proposed \textsc{\textbf{StereoFlow}} achieves SoTA performance on the Scene Flow test set, surpassing RAFT-Stereo~\cite{raftstereo} by $42\%$ and DEFOM-Stereo~\cite{defomstereo} by $23\%$ on EPE.
For zero-shot generalization evaluation in Tab.~\ref{tab:zero-shot}, we directly employ the synthetic pre-trained model to realistic datasets, including KITTI, Middlebury and ETH3D, without any domain-specific fine-tuning. 
In this setting, the proposed \textsc{\textbf{StereoFlow}} again achieves SoTA performances, reducing the error rate over DEFOM-Stereo~\cite{defomstereo} by $6\%$ on KITTI-2012, by $20\%$ on KITTI-2015, by $27\%$ on Middlebury-Quarter and by $26\%$ on ETH3D.
The qualitative comparisons in Fig.~\ref{fig:intro} demonstrate that the proposed \textsc{\textbf{StereoFlow}} achieves strong geometric consistency and fine-grained details, even under zero-shot generalization.

\noindent \textbf{Realistic Benchmark Evaluation.}
\label{sec:realistic}
\setlength{\tabcolsep}{5pt}
\begin{table}[t]
    \centering
    \caption{\textbf{Evaluation of reflective regions on KITTI-2012.} All metrics are reported as percentages. }
    \vspace{-5pt}
    \resizebox{0.48\textwidth}{!}{
    \begin{tabular}{l|cccccc}
    \toprule
    \multirow{3}{*}{\textbf{Methods}} & \multicolumn{6}{c}{\textbf{KITTI-2012 Reflective Region}} \\
    \cline{2-7}
    & Out-2 & Out-2 & Out-3 & Out-3 & Out-4 & Out-4\\
    & Noc & All & Noc & All & Noc & All  \\
    \midrule
    ACVNet~\cite{acvnet} & 11.42 & 13.53 & 7.03 & 8.67 & 5.18 & 6.48\\
    CREStereo~\cite{crestereo} & 9.71 & 11.26 & 6.27 & 7.27 & 4.93 & 5.55\\
    IGEV~\cite{igev} & 7.57 & 8.80 & 4.35 & 5.00 & 3.16 & 3.57 \\
    Selective-IGEV~\cite{selective} & 6.73 & 7.84 & 3.79 & 4.38 & 2.66 & 3.05  \\
    LoS~\cite{los} & 6.31 & 7.84 & 3.47 & 4.45 & 2.41 & 3.01 \\
    NMRF~\cite{nmrf} & 10.02 & 12.34 & 6.35 & 8.11 & 4.80 & 6.09\\
    DEFOM-Stereo~\cite{defomstereo} & 5.76 & \underline{6.72} & 3.04 & 3.56 & 1.93 & 2.27\\
    BridgeDepth~\cite{bridgedepth} & 5.80 & 6.85 & 2.91 & 3.48 & 1.88 & 2.23\\
    MonSter~\cite{monster} & \underline{5.66} & 6.81 & \underline{2.75} & \underline{3.38} & \textbf{1.73} & \underline{2.13}\\
    \midrule
    \textbf{\textsc{StereoFlow}} (Ours) & \textbf{5.62} & \textbf{6.70} & \textbf{2.71} & \textbf{3.29} & \underline{1.74} & \textbf{2.11}\\
    \bottomrule
    \end{tabular}
    }
    \vspace{-8pt}
    \label{tab:reflective}
\end{table}

\subsubsection{\textbf{ETH3D}}
We finetune the synthetic pretrained \textsc{\textbf{StereoFlow}} for 300k steps on a mixture of the Basic Training Set and ETH3D~\cite{eth3d}.
As shown in Tab.~\ref{tab:mix}, the proposed model obtains the top performance on the ETH3D benchmark and ranks $1^{st}$ in multiple metrics.
Specifically, the proposed model surpasses DEFOM-Stereo~\cite{defomstereo} by $17\%$ on Bad 1.0 (noc) and $12\%$ on Bad 1.0 (all).
The qualitative comparisons in Fig.~\ref{fig:exp} demonstrate that the proposed \textsc{\textbf{StereoFlow}} achieves strong geometric consistency and fine-grained details across challenging ill-posed regions, including reflective surfaces, highlights, textureless, occluded and discontinuous areas.    

\subsubsection{\textbf{Middlebury}}
We finetune the synthetic pre-trained \textsc{\textbf{StereoFlow}} for 200k steps on a mixture of the Basic Training Set and Middlebury~\cite{middlebury14}.
As shown in Tab.~\ref{tab:mix}, the proposed model obtains the top performance on the Middlebury benckmark and ranks $1^{st}$ in multiple metrics.
Specifically, the proposed model surpasses DEFOM-Stereo~\cite{defomstereo} by $6\%$ on Bad 2.0 (noc) and $6\%$ on Bad 2.0 (all).
The qualitative comparisons in Fig.~\ref{fig:exp} demonstrate that the proposed \textsc{\textbf{StereoFlow}} achieves strong geometric consistency and fine-grained details across challenging ill-posed regions, including textureless, occluded and discontinuous areas.  

\subsubsection{\textbf{KITTI}}
We finetune the synthetic pre-trained \textsc{\textbf{StereoFlow}} for 50k steps on a mixture of the KITTI-2012~\cite{kitti12}, KITTI-2015~\cite{kitti15} and Virtual KITTI-2~\cite{kitti2}.
As shown in Tab.~\ref{tab:mix}, the proposed model obtains the top performance on the KITTI-2012 and KITTI-2015 benchmarks and ranks $1^{st}$ in multiple metrics.
On KITTI-2015, the proposed model surpasses DEFOM-Stereo~\cite{defomstereo} by $10\%$ on D1-bg (noc) and $10\%$ on D1-bg (all).
On KITTI-2012, the proposed model surpasses DEFOM-Stereo~\cite{defomstereo} by $8\%$ on Out-2 (noc) and $5\%$ on Out-2 (all).

\subsubsection{\textbf{KITTI Reflective Region}}
As shown in Tab.~\ref{tab:reflective}, \textsc{\textbf{StereoFlow}} ranks $1^{st}$ on the KITTI-2012 benckmark for multiple metrics of reflective regions.
Specifically, the proposed model surpasses BridgeDepth~\cite{bridgedepth} by $4\%$ on Out-2 (noc), by $7\%$ on Out-3 (noc) and by $8\%$ on Out-4 (noc). 
The quantitative comparisons show that the proposed \textsc{\textbf{StereoFlow}} a significant improvement in realistic datasets and ill-posed regions.

\setlength{\tabcolsep}{16pt}
\begin{table*}[]
    \caption{\textbf{Flow Matching Comparisons.} 
    Probability path definitions for existing methods which fit in the diffusion paths (top), independent coupling path (middle) and optimal coupling path (bottom). We define a new optimal coupling path objective (termed Transition Flow Matching) that can handle general source distributions and optimal transport flows.
    }
    \vspace{-6pt}
    \centering
    \resizebox{0.98\textwidth}{!}{
    \renewcommand{\arraystretch}{0.5}
    \begin{tabular}{l|lll}
    \toprule
    \textbf{Probability Path} & $\boldsymbol{p(x)}$ & $\boldsymbol{\mu_t}$ & $\boldsymbol{\sigma_t}$ \\
    \hline
    \rowcolor[gray]{0.85}\multicolumn{4}{l}{\textbf{Diffusion Path}} \\
    Variance Exploding (VE)~\cite{ddpm}  & $p(x_1)$ & $x_1$ & $\sigma_{1 - t}$ \\
    Variance Preserving (VP)~\cite{smld} & $p(x_1)$ & $\alpha_{1-t} x_1$ & $\sqrt{1 - \alpha_{1-t}^2}$ \\
    Conditional Flow Matching (CFM)~\cite{fm} & $p(x_1)$ & $tx_1$ & $ 1 - (1 - \sigma_{min}) t  $ \\
    \hline
    \rowcolor[gray]{0.85}\multicolumn{4}{l}{\textbf{Independent Coupling Path}} \\
    Rectified Flow~\cite{reflow} & $p(x_0) p(x_1)$ & $t x_1 + (1-t) x_0$ & $0$  \\
    Independent CFM~\cite{otfm} & $p(x_0) p(x_1)$ & $t x_1 + (1-t) x_0$ & $\sigma_{min}$ \\
    VP Stochastic Interpolant~\cite{stochastic} & $p(x_0) p(x_1)$ & $\cos(\frac{\pi}{2} t) x_0 + \sin(\frac{\pi}{2} t) x_1$ & $0$ \\
    \hline
    \rowcolor[gray]{0.85}\multicolumn{4}{l}{\textbf{Optimal Coupling Path}} \\
    Optimal Transport CFM~\cite{otfm} & $\pi(x_0, x_1)$ & $t x_1 + (1-t) x_0$ & $\sigma_{min}$ \\
    Schrödinger Bridge CFM~\cite{otfm} & $\pi(x_0, x_1)$ & $t x_1 + (1-t) x_0$ & $\smash{\sigma \sqrt{t(1-t)}}$  \\
    Trigonometric Interpolant~\cite{stochastic}  & $\pi(x_0, x_1)$ & $\cos(\frac{\pi}{2} t) x_0 + \sin(\frac{\pi}{2} t) x_1$ & $\smash{\sigma \sqrt{t (1-t)}}$ \\
    Gaussian Encoding-Decoding~\cite{stochastic} & $\pi(x_0, x_1)$ & $\cos^2(\pi t) \, 1_{[0,\frac{1}{2})}(t) x_0 + \cos^2(\pi t) \, 1_{(\frac{1}{2},1]}(t) x_1$ & $\sin^2(\pi t)$ \\
    \textbf{Transition Flow Matching (Ours)}  & $\pi(x_0, x_1)$ & $t x_1 + (1-t) x_0$ & $\sigma_{min} t + \sigma_{max} (1-t)$ \\
    \bottomrule
    \end{tabular}
    }
    \vspace{-2pt}
    \label{tab:pp}
\end{table*}
\setlength{\tabcolsep}{12pt}
\begin{table}[t]
    \centering
    \caption{\textbf{Configurations of StereoDiT.} The subscripts indicate the downsampling factor applied to the full resolution.}
    \vspace{-5pt}
    \resizebox{0.42\textwidth}{!}{
    \begin{tabular}{l|cc}
    \toprule
    & \textbf{StereoDiT}$_{1/8}$ & \textbf{StereoDiT}$_{1/4}$ \\
    \hline
    \rowcolor[gray]{0.9}\multicolumn{3}{l}{\textbf{Base}} \\
    params & 12M & 20M  \\
    \hline
    \rowcolor[gray]{0.9}\multicolumn{3}{l}{\textbf{Encoder}} \\
    patch size & \multicolumn{2}{c}{16} \\
    depth & 4 & 8  \\
    REPAlign layer & 2 & 4  \\
    hidden dim & \multicolumn{2}{c}{256}  \\
    heads & \multicolumn{2}{c}{8}  \\
    \hline
    \rowcolor[gray]{0.9}\multicolumn{3}{l}{\textbf{Decoder}} \\
    patch size & \multicolumn{2}{c}{1} \\
    depth & 2 & 3 \\
    hidden dim & \multicolumn{2}{c}{32} \\
    \hline
    \rowcolor[gray]{0.9}\multicolumn{3}{l}{\textbf{Training}} \\
    optimizer & \multicolumn{2}{c}{AdamW, $\beta_1, \beta_2=0.9, 0.999$} \\
    learning rate & \multicolumn{2}{c}{1e-4} \\ 
    weight decay & \multicolumn{2}{c}{0} \\ 
    ema decay & \multicolumn{2}{c}{0.999} \\
    time sampler & \multicolumn{2}{c}{$t\sim \text{U}[0,1]$} \\
    \hline
    \rowcolor[gray]{0.9}\multicolumn{3}{l}{\textbf{Sampling}} \\
    ODE solver & \multicolumn{2}{c}{Euler} \\
    time steps & \multicolumn{2}{c}{linear in [0.0, 1.0]} \\
    ODE steps & 2 & 2 \\
    sampling time & 0.04s & 0.07s  \\
    \bottomrule
    \end{tabular}
    }
    \vspace{-8pt}
    \label{tab:config}
\end{table}
\subsection{Configurations of StereoDiT}
\label{sec:configurations}
As shown in Tab.~\ref{tab:config}, we summarize the configurations of StereoDiT$_{1/8}$ and StereoDiT$_{1/4}$. 
The proposed architecture follows the conventions of recent generative transformers, such as DiT~\cite{dit}, SiT~\cite{sit}, PixNerd~\cite{pixnerd}, and PixDiT~\cite{pixeldit}.
However, unlike these general-purpose generators, StereoDiT achieves a remarkably more compact architecture without compromising performance, which is enabled by three task-specific innovations:
(i) a prior-guided formulation that significantly narrows the generative solution space;
(ii) a Transition Flow Matching objective that promotes efficient learning dynamics and few-step sampling; 
(iii) a frequency-decoupled transformer architecture tailored specifically for modeling correspondence ambiguity.

\subsection{Comparisons of Flow Matching Objective}
As shown in Tab.~\ref{tab:pp}, we summarize the definitions of the flow matching objectives for three distinct path configurations:
(i) diffusion paths—where one endpoint is sampled from an isotropic gaussian distribution; 
(ii) independent coupling paths—where both endpoints are sampled independently from the data distribution; 
and (iii) optimal coupling paths—where both endpoints are sampled jointly from the data distribution.
To combine the complementary strengths of diffusion paths and optimal paths, we intrudoce Transition Flow Matching, an optimal coupling-based flow matching with a linear interpolant for directional displacement of geometric structures and a linear noise scheduler to synthesize details.
By jointly exploiting deterministic transport and stochastic perturbations, this objective preserves the disparity priors inherent in stereo matching, enabling efficient learning dynamics and optimization.

\subsection{Ablation Study} 
\label{sec:ablation}
\setlength{\tabcolsep}{10pt}
\begin{table*}[t]
    \caption{\textbf{Ablation study on Scene Flow test set and KITTI.} The baseline is a simplified DEFOM-Stereo~\cite{defomstereo}. The inference time is evaulated with the input of 960$\times$540 on an NVIDIA 4090 GPU.}
    \vspace{-6pt}
    \centering
    \resizebox{1\textwidth}{!}{
    \begin{tabular}{l|cccc|c|c|c|c}
    \toprule
    \multirow{2}{*}{\textbf{Methods}} & \multicolumn{4}{c|}{\textbf{Proposed Modules}} & \multicolumn{1}{c|}{\textbf{Scene Flow}} & \multicolumn{1}{c|}{\textbf{KITTI-12}} & \multicolumn{1}{c|}{\textbf{KITTI-15}} & \multirow{2}{*}{\textbf{Time (s)}} \\ 
    & Cascade & Progressive & StereoDiT$_{1/8}$ & StereoDiT$_{1/4}$ & EPE & Bad 3.0 & Bad 3.0 \\
    \midrule
    Baseline & & & & & 0.43 & 4.14 & 5.27 & 0.29 \\ 
    \midrule
    + Cascade & \checkmark & & & & 0.41 & 3.91 & 4.97 & 0.26 \\
    + Progressive & & \checkmark & & & 0.37 & 3.80 & 4.55 & 0.28 \\
    + StereoDiT$_{1/8}$ & & \checkmark & \checkmark & & 0.35 & 3.63 & 4.26 & 0.32 \\ 
    + StereoDiT$_{1/4}$ & & \checkmark & & \checkmark & 0.34 & 3.61 & 4.19 & 0.35 \\
    \midrule
    Full Model  & & \checkmark & \checkmark & \checkmark & 0.32 & 3.55 & 3.92 & 0.39 \\ 
    \bottomrule
    \end{tabular}
    }
    \vspace{-2pt}
    \label{tab:abla}
\end{table*}
\begin{figure*}[!t]
    \centering
    \includegraphics[width=1\textwidth]{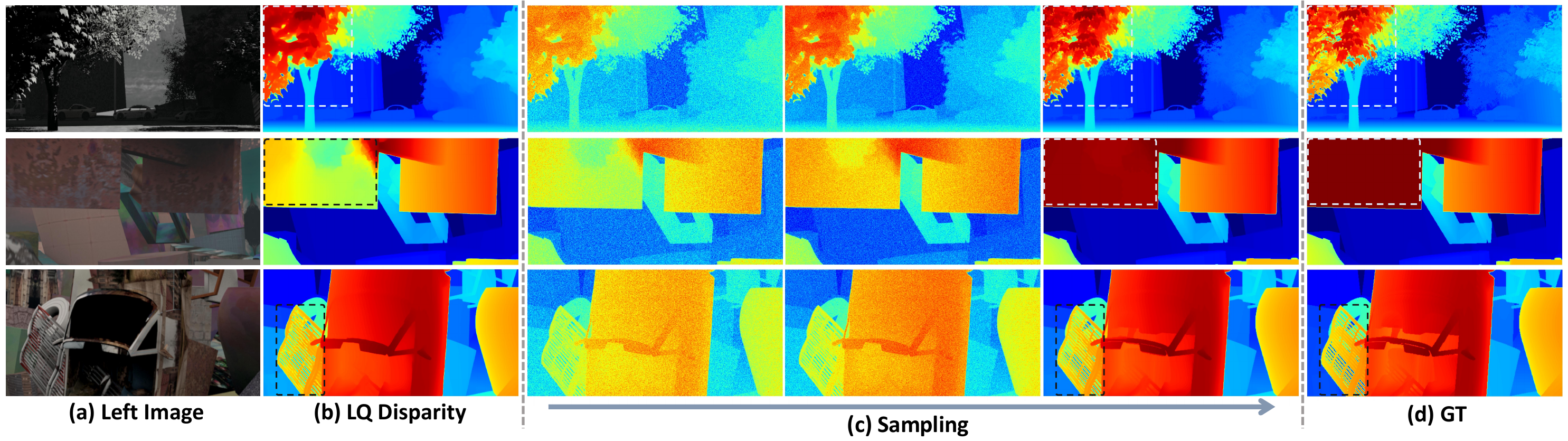}
    \vspace{-17pt}
    \caption{
    \textbf{Sampling Visualization of Transition Flow Matching.} 
    Starting directly from the noisy degraded disparities, with only two sampling steps, the proposed Transition Flow Matching enables the displacement of geometric structures and the synthesis of high-frequency details, with improvements in challenging ill-posed regions, such as reflective surfaces, textureless, occluded and discontinuous areas.
    (\textit{Zoom in for a better view.})
    }
\label{fig:tfm}
\end{figure*}
To evaluate the impact of components within the proposed \textsc{\textbf{StereoFlow}}, we conduct several ablation studies. 

\subsubsection{\textbf{Effectiveness of progressive cascade matching network}}
Our baseline is a simplified version of DEFOM-Stereo~\cite{defomstereo}, with its affine-invariant monocular depth and metric scaling modules removed.
As shown in Tab.~\ref{tab:abla}, against this baseline, the cascade matching network with a 3D ACV at each stage brings marginal improvements.
More notably, the progressive cascade matching network with a low-resolution 4D GEV for geometric consistency and a high-resolution 3D ACV for fine-grained details yields substantial improvements.
Steady improvements are also observed under zero-shot generalization.

\subsubsection{\textbf{Effectiveness of StereoDiT}}
To model the correspondence ambiguity unresolved by deterministic matching networks, we introduce StereoDiT, a diffusion transformer tailored for stereo matching.
As shown in Tab.~\ref{tab:abla}, integrating StereoDiT at either resolution improves performance, and integrating it at both the two resolutions delivers greater improvements.
Thanks to our optimizations, the system delivers a competitive inference time of 0.39s despite integrating two diffusion models, outperforming 0.64s for MonSter~\cite{monster} and remaining competitive with 0.35s for IGEV-Stereo~\cite{igev}.

\subsubsection{\textbf{Effectiveness of Transition Flow Matching}}
We introduce Transition Flow Matching, a few-step flow matching objective that replaces isotropic gaussian priors with disparity priors.
As illustrated in Fig.~\ref{fig:tfm}, with only two sampling steps, both geometric structures and details are enhanced, along with improvements in challenging ill-posed regions, such as reflective surfaces, textureless, occluded and discontinuous areas.

\subsubsection{\textbf{Ablation study about GRU iterations}}
\setlength{\tabcolsep}{8pt}
\begin{table}[t]
    \centering
    \caption{\textbf{Ablation Study of GRU Iterations on Scene Flow test set.} The subscripts indicate the downsampling factor applied to the full resolution. The metric used for comparison is EPE.} 
    \vspace{-5pt}
    \resizebox{0.48\textwidth}{!}{
    \begin{tabular}{c|ccccc}
    \toprule
    \diagbox{\textbf{Iter. of GRU}$_{1/8}$}{\textbf{Iter. of GRU}$_{1/4}$} & 1 & 2 & 4 & 6 & 8 \\
    \midrule
    1 &0.59 &0.54 &0.50 &0.45 &0.44  \\
    2 &0.54 &0.52 &0.48 &0.43 &0.41  \\
    4 &0.52 &0.50 &0.46 &0.41 &0.39  \\
    6 &0.48 &0.47 &0.42 &0.40 &0.38  \\
    8 &0.47 &0.43 &0.41 &0.39 &\textbf{0.37}  \\
    \bottomrule
    \end{tabular}
    }
    \vspace{-8pt}
    \label{tab:gru_iters}
\end{table}
As shown in Tab.~\ref{tab:gru_iters}, the proposed progressive cascade matching network achieves superior performance with only 16 total iterations (8+8), substantially fewer than the 32 iterations required by baseline methods such as IGEV-Stereo~\cite{igev} and DEFOM-Stereo~\cite{defomstereo}.
This efficiency stems from the strategy of low-resolution consistent disparity reconstruction and high-resolution fine-grained disparity recovery.

\subsubsection{\textbf{Ablation study about StereoDiT}}
\setlength{\tabcolsep}{9pt}
\begin{table}[t]
    \centering
    \caption{\textbf{Ablation Study of StereoDiT on Scene Flow test set.} The evaluation is based on the proposed Transition Flow Matching objective. The baseline is PixelDiT~\cite{dit}. The metric used for comparison is EPE.}
    \resizebox{0.48\textwidth}{!}{
    \begin{tabular}{l|cccccc}
    \toprule
    \multirow{2}{*}{\textbf{Model}} &\multicolumn{5}{c}{\textbf{Number of Sampling}} \\
    \cline{2-6}
    & 1 & 2 & 3 & 4 & 5 \\
    \midrule
    Baseline (PixelDiT) &0.527 &0.419 &0.380 &0.362 &0.356   \\
    + Decoupled arctitecture &0.499 &0.391 &0.361 &0.349 &0.342 \\
    + REPAlign &0.463 &0.380 &0.368 &0.351 &0.349 \\
    + Local cost volume &0.458 &0.358 &0.352 &0.351 &0.351 \\
    + Uncertainty &0.450 &\textbf{0.348} &0.348 &0.348 &0.347 \\
    \bottomrule
    \end{tabular}
    }
    \vspace{-8pt}
    \label{tab:stereodit}
\end{table}
\setlength{\tabcolsep}{12pt}
The proposed StereoDiT is a pixel diffusion transformer with a decoupled architecture comprising a geometric encoder and a stereo decoder. 
The geometric encoder employs monocular depth features to enforce low-frequency geometric consistency, while the stereo decoder employs a set of specialized stereo conditions to recover high-frequency metric details.
The ablation study in Tab.~\ref{tab:stereodit} validates that the proposed designs consistently improve performance while simultaneously decreasing sampling steps, ultimately achieving substantial improvements with only two sampling steps.

\section{Conclusion \& Limitation}
\label{sec:conclusion}
In this paper, we introduce a prior-guided generative framework that integrates deterministic matching regression and generative distribution modeling within a complementary formulation.
Built upon this formulation, we propose \textsc{\textbf{StereoFlow}} through three key components:
(i) a two-stage progressive cascade matching network that progressively produces multi-resolution stereo conditions with complementary matching cues;
(ii) a pixel diffusion transformer (termed StereoDiT) with a frequency-decoupled architecture for modeling correspondence ambiguity; 
(iii) a few-step flow matching objective (termed Transition Flow Matching) for efficient optimization and few-step sampling.
In summary, we achieves strong geometric consistency and fine-grained details in ill-posed, discontinuous regions and under zero-shot generalization.

The proposed model has two limitations: 
(i) although the proposed designs have shortened the sampling time, diffusion models still inevitably introduce non-negligible latency, and one-step sampling strategy could be introduced in the future to mitigate this issue;
(ii) the two-stage training strategy not only increases complexity but also introduces instability, highlighting the need for a more robust end-to-end one-stage training approach in the future.

\footnotesize
\bibliographystyle{IEEEtran}
\bibliography{arxiv}

}
\end{document}